\pdfoutput=1

\documentclass[11pt]{article}

\usepackage{ACL2023}

\usepackage{times}
\usepackage{latexsym}

\usepackage[T1]{fontenc}

\usepackage[utf8]{inputenc}

\usepackage{microtype}

\usepackage{inconsolata}

\usepackage{graphicx}
\usepackage{multirow}
\usepackage{subcaption}
\usepackage{enumitem}
\usepackage{booktabs}
\usepackage{wrapfig}
\usepackage{amsmath}
\usepackage{xcolor}
\usepackage{hyperref}

\usepackage[compact]{titlesec}
\titlespacing{\section}{0pt}{2ex}{1ex}
\titlespacing{\subsection}{0pt}{1ex}{0ex}
\titlespacing{\subsubsection}{0pt}{0.5ex}{0ex}

\newcommand{\data}[1]{\textit{{#1}}}
\newcommand{\header}[1]{\textit{{#1}}}
\newcommand{\type}[1]{\uppercase{#1}}
\newcommand{\method}[1]{\texttt{{#1}}}
\newcommand{\deptool}[1]{\texttt{{#1}}}

\newcommand{\spandetect}{\textit{Span Detection Model}}
\newcommand{\spanclass}{\textit{Span Classification Model}}

\newcommand{\splitner}{\method{SplitNER}}
\newcommand{\our}{\method{SplitNER(QA-QA)}}
\newcommand{\spanseq}{\method{SplitNER(SeqTag-QA)}}
\newcommand{\nocharpattern}{\method{SplitNER(QA$_{NoCharPattern}$-QA)}}
\newcommand{\bertqa}{\method{Single(QA)}}
\newcommand{\bertseq}{\method{Single(SeqTag)}}

\newcommand{\securitydata}{\data{CTIReports}}

\newcommand{\bertlarge}{BERT-\textit{large}}

\title{Split-NER: Named Entity Recognition via Two Question-Answering-based Classifications}

\author{Jatin Arora \\
  Nuro Inc. \\
  \texttt{jarora@nuro.ai} \\\And
  Youngja Park \\
  IBM T.J. Watson Research Center\\
  \texttt{young\_park@us.ibm.com} \\}

\begin{document}

\maketitle
\begin{abstract}
In this work, we address the NER problem by splitting it into two logical sub-tasks: (1) \textit{Span Detection} which simply extracts mention spans of entities, irrespective of entity type; 
(2) \textit{Span Classification} which classifies the spans into their entity types. 
Further, we formulate both sub-tasks as question-answering (QA) problems 
and produce two leaner models which can be optimized separately for each sub-task.
Experiments with four cross-domain datasets demonstrate that this two-step approach is both effective and time efficient. 
Our system, \splitner{} outperforms baselines on \data{OntoNotes5.0}, \data{WNUT17} and a cybersecurity dataset and gives on-par performance on \data{BioNLP13CG}. 
In all cases, it achieves a significant reduction in training time compared to its QA baseline counterpart.
The effectiveness of our system stems from fine-tuning the BERT model twice, separately for span detection and classification. The source code can be found at \texttt{\href{https://github.com/c3sr/split-ner}{github.com/c3sr/split-ner}}. 



\end{abstract}

\section{Introduction}
\label{sec:intro}
Named entity recognition (NER) is a foundational task for a variety of applications like question answering and machine translation~\cite{li2020survey}. 
Traditionally, NER has been seen as a sequence labeling task where a model is trained to classify each token of a sequence to a predefined  class 
~\cite{carreras-etal-2002, carreras-etal-2003, Chiu16,Lample16,ma2016end,devlin2019bert,WanR0022}.

Recently, there has been a new trend of formulating NER as span prediction problem~\cite{Stratos17b,li2020MRC,Jiang20,Ouchi20,fu2021spanner},
where a model is trained to jointly perform span boundary detection and multi-class classification over the spans. 
Another trend is to formulate NER as a question answering (QA) task~\cite{li2020MRC}, where the model is given a sentence and a query corresponding to each entity type. 
The model is trained to understand  the query and extracts mentions of the entity type  as answers. 
While these new frameworks have shown improved results, both approaches suffer from a high computational cost: span-based NER systems consider all possible spans (i.e., $n^2$ (quadratic) spans for a sentence with $n$ tokens) and the QA-based system multiplies each input sequence by the number of entity types resulting in  \textit{N}$\times$\textit{T} input sequences for \textit{N} sentences and $\textit{T}$ entity types.

In this work, we borrow the effectiveness of span-based and QA-based techniques and  make it more efficient by breaking (splitting up) the NER task into a two-step pipeline of classification tasks. In essence, our overall approach comes under the span-based NER paradigm, and each sub-task is formulated as a QA task inspired by the higher accuracy offered by the QA framework.
The first step, \textit{Span Detection} performs token-level classification to extract mention spans from text, irrespective of entity type and 
the second step, \textit{Span Classification} classifies the extracted spans into their corresponding entity type, thus completing the NER task. 
Unlike other span-based NER techniques which are quadratic in terms of sequence length, our \textit{Span Detection} process is linear.
Compared to other QA-based techniques which query for all entity types in each sentence, our \textit{Span Classification}  queries each sentence only once for each entity mention in the sentence. This makes it highly efficient for datasets with large number of entity types like \textit{OntoNotes5.0}. 

%

\if false
The main contributions of this work include:
\begin{itemize}[topsep=2pt,itemsep=1pt,parsep=1pt]
 \item We show that entity span detection and type classification are not tightly coupled tasks and can be handled independently. 
\item  In fact, the separation helps to produces a more effective NER system, since we can fine-tune the models for each sub-task by applying
different features, loss functions and model architectures (e.g., LSTM for span detection and BERT for span classification).   

\item We demonstrate that, in a QA-based setup, this proposed logical separation is more efficient than modeling NER as a single atomic task and takes much lesser time to train with a comparable or even better performance.

\item Further, our system can handle structured data, where the entity spans are already known. In this case, we can run only the entity type classification model.  

\end{itemize}
\fi


\section{Method}
\label{sec:method}
\begin{figure*}[h!]
    \centering
    \includegraphics[width=0.86\linewidth]{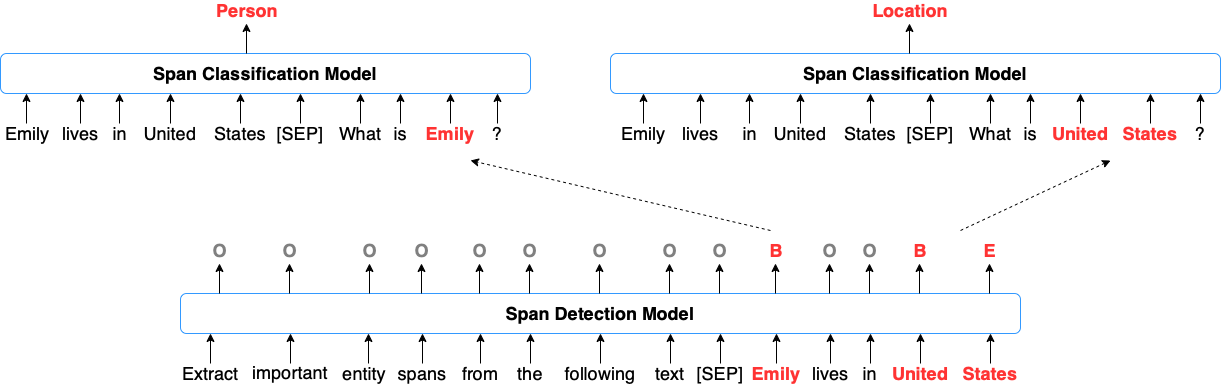}
\setlength{\belowcaptionskip}{-12pt}
    \caption{\splitner{} System Overview. \spandetect{} identities two mention spans (\textit{Emily} and \textit{United States}) from the input (\textit{``Emily lives in United States"}), 
    and \spanclass{} assigns each span to their entity type.} 
    \label{fig:framework}
\end{figure*}

Figure~\ref{fig:framework} illustrates how our two-step \splitner{} system works. 
\spandetect{} is entity-agnostic and identifies all mention spans irrespective of entity type. 
The extracted spans are passed to \spanclass{} which reanalyses them in the sentence structure and classifies them into an entity type. 
Both models use BERT-\textit{base} as their underlying architecture and are designed as QA tasks. Hence, moving forward, we may sometimes explicitly call our system as \our{} to distinguish it from other variants we experiment with.

\subsection{Span Detection}
\label{sec:span}
Given a sentence $\mathcal{S}$ as a $n$-length sequence of tokens, $\mathcal{S} = \langle w_1, w_2 \ldots w_n \rangle$, the goal is to output a list of spans  $\langle s, e\rangle$, where $s, e \in [1, n]$ are \textit{start} and \textit{end} indices of a mention. 
We formulate this as a QA task 
classifying each token using \texttt{BIOE} scheme\footnote{All experiments in this paper use \texttt{BIOE} scheme but the approach is generalizable to other schemes like \texttt{BIOES}.}. 
Since the goal is to detect spans irrespective of their entity type, we use a generic question, ``\textit{Extract important entity spans from the following text}'', prefixed with input sentence (see Figure \ref{fig:framework})\footnote{Other similar question texts / no question text also gives similar results as shown in ablation study in Appendix \ref{sec:ablations}.}. 

\if false
\begin{figure*}[htp]
\centering
\begin{subfigure}[b]{0.35\textwidth}
   \includegraphics[width=\linewidth]{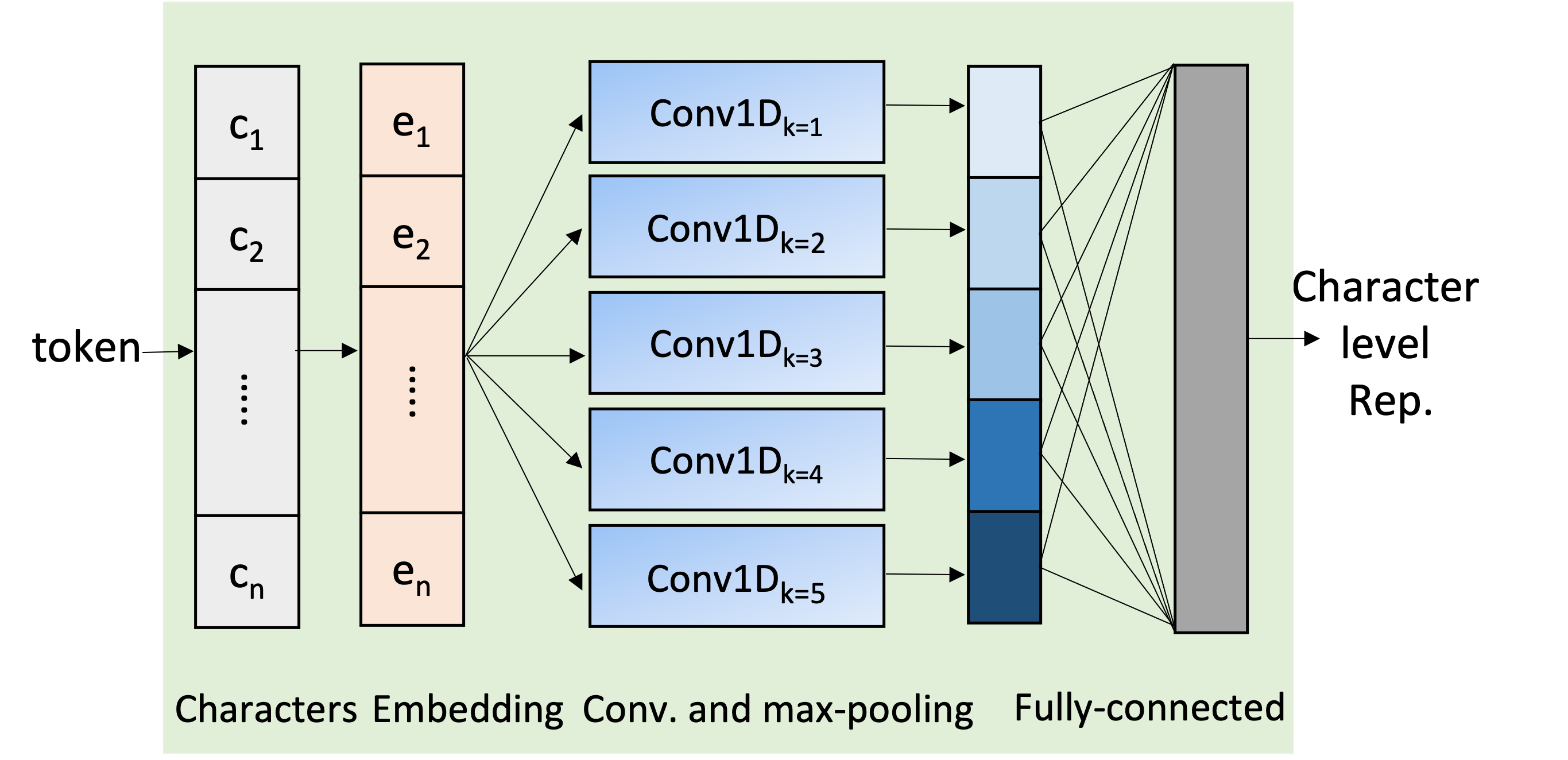}
   \caption{Character Sequence Feature Learning}
   \label{fig:char-feature}
\end{subfigure}
\begin{subfigure}[b]{0.33\textwidth}
   \hspace{-5mm}\includegraphics[width=\linewidth]{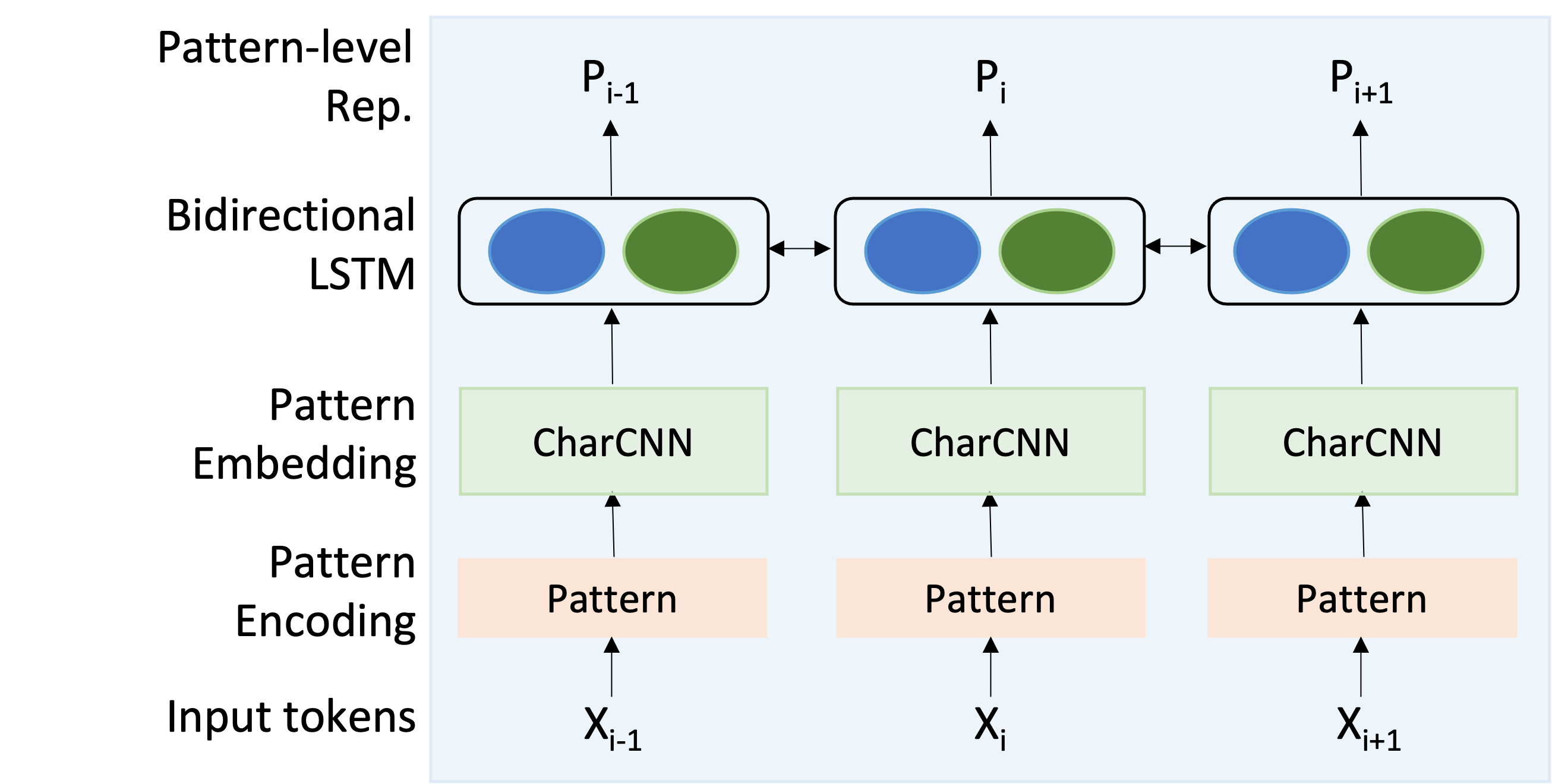}
   \caption{Orthographic Feature Learning}
   \label{fig:pattern-feature}
\end{subfigure}
\begin{subfigure}[b]{0.3\textwidth}
   \includegraphics[width=\linewidth]{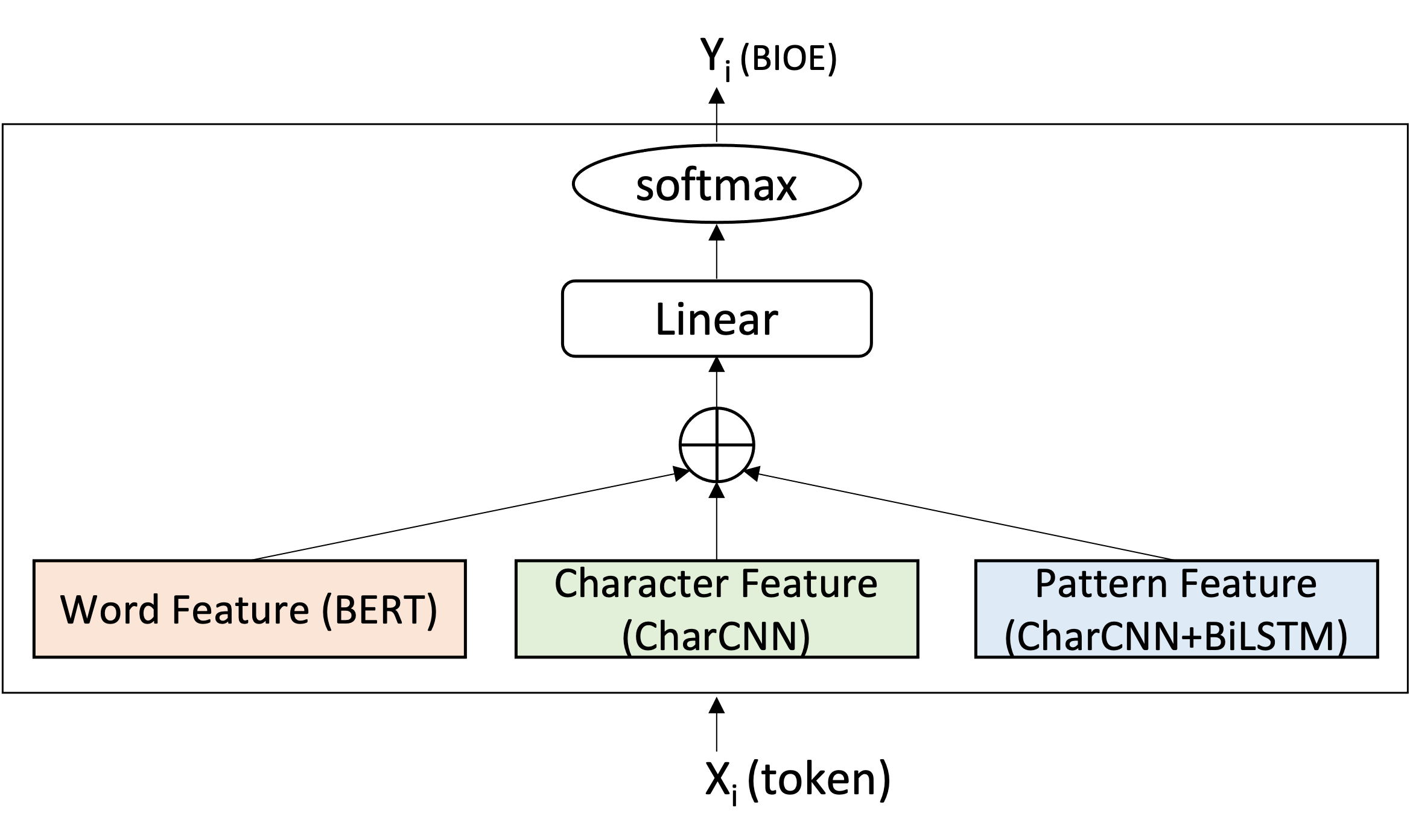}
   \caption{Token-level Schematic Model}
   \label{fig:token-span}
\end{subfigure}
   \caption{Architecture for \spandetect{}}
   \label{fig:span-detection}
\end{figure*}
\fi

\begin{figure*}[htp]
\centering
\begin{subfigure}[b]{0.45\textwidth}
   \includegraphics[width=\linewidth]{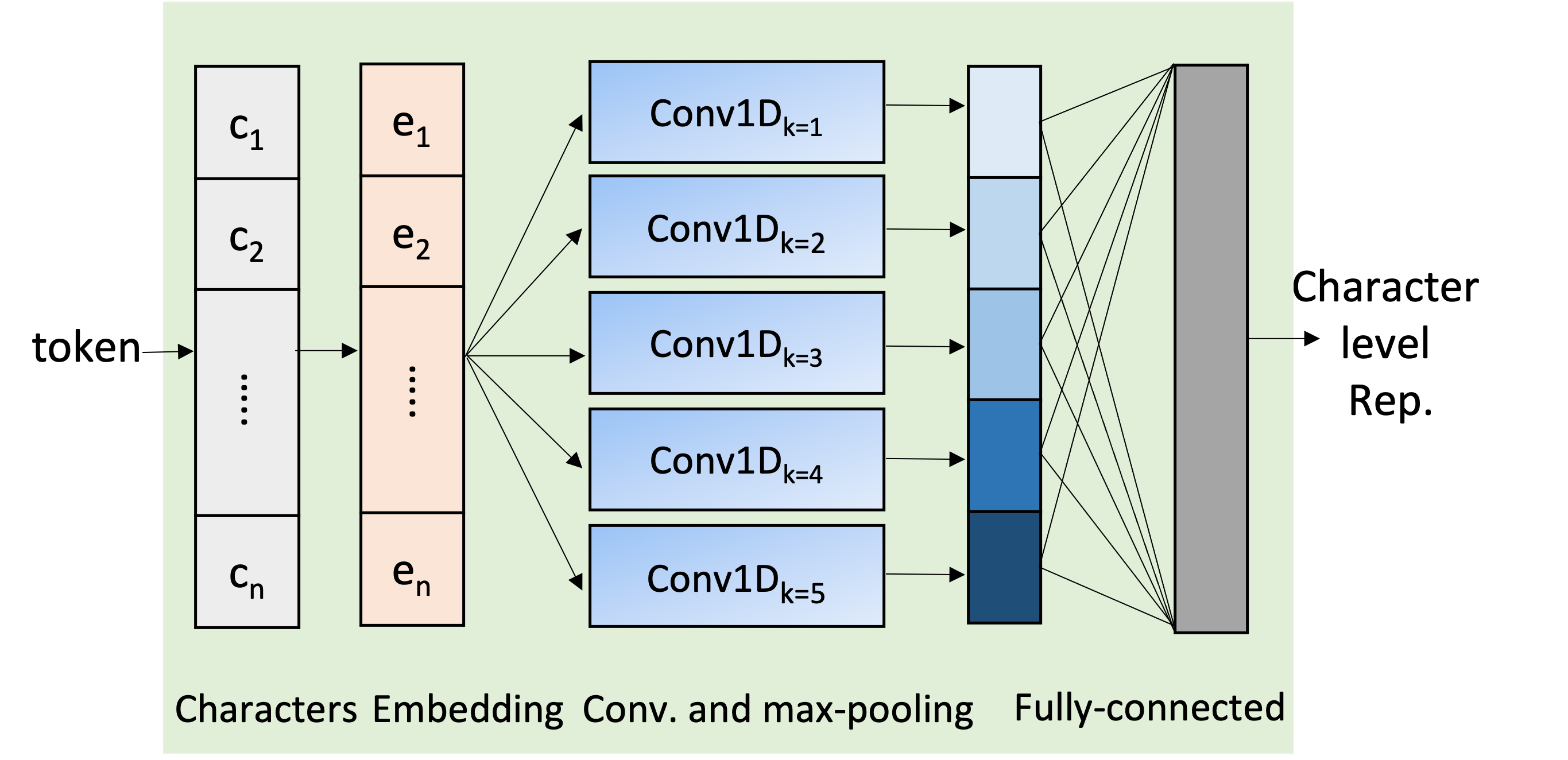}
   \caption{Character Sequence Feature Learning}
   \label{fig:char-feature}
\end{subfigure}
\begin{subfigure}[b]{0.45\textwidth}
   \hspace{-5mm}\includegraphics[width=\linewidth]{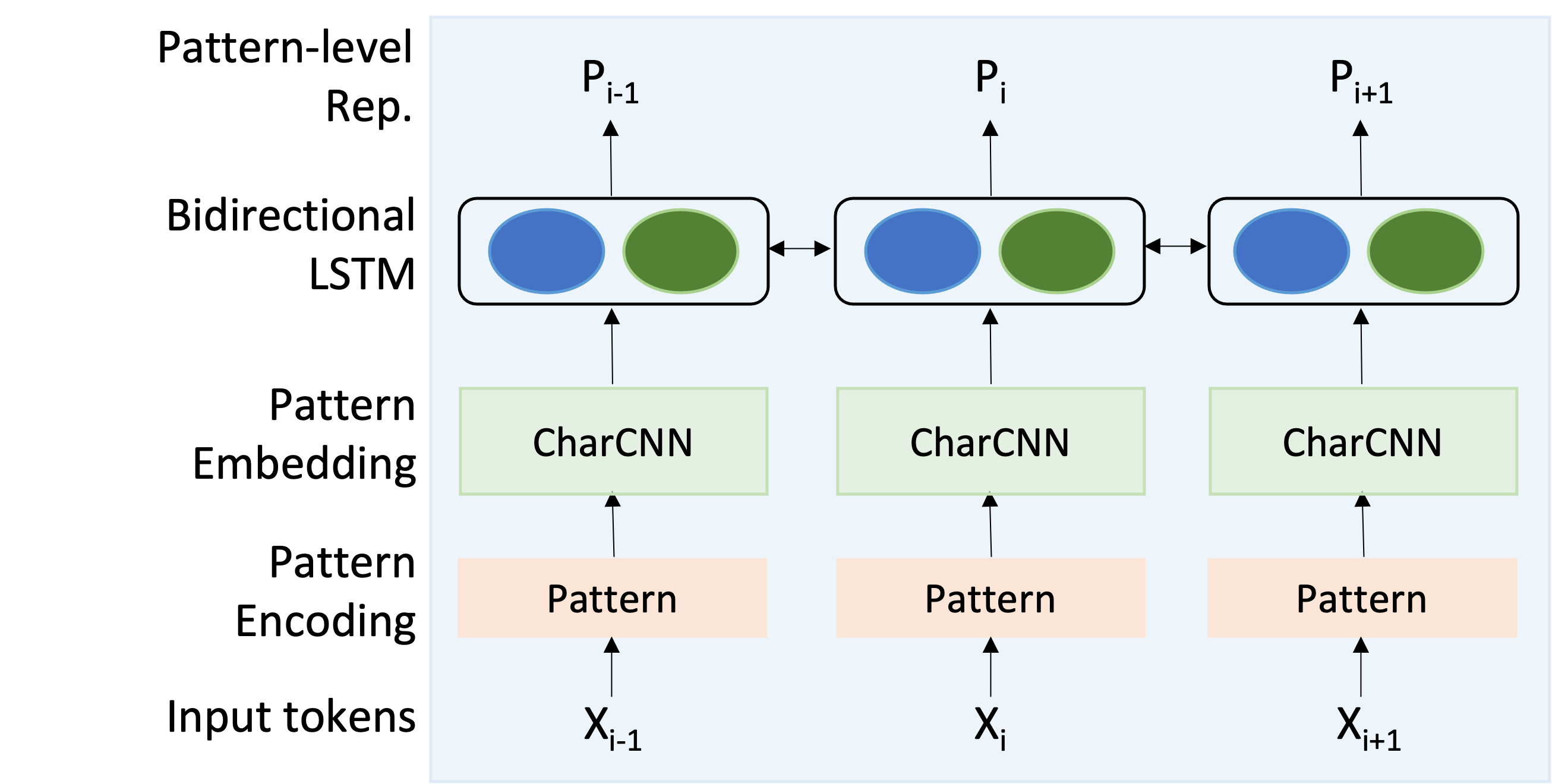}
   \caption{Orthographic Feature Learning}
   \label{fig:pattern-feature}
\end{subfigure}
   \caption{Architecture for Character and Pattern Feature Learning}
   \label{fig:char-pattern}
\end{figure*}

\begin{figure}[htp]
   \includegraphics[width=\linewidth]{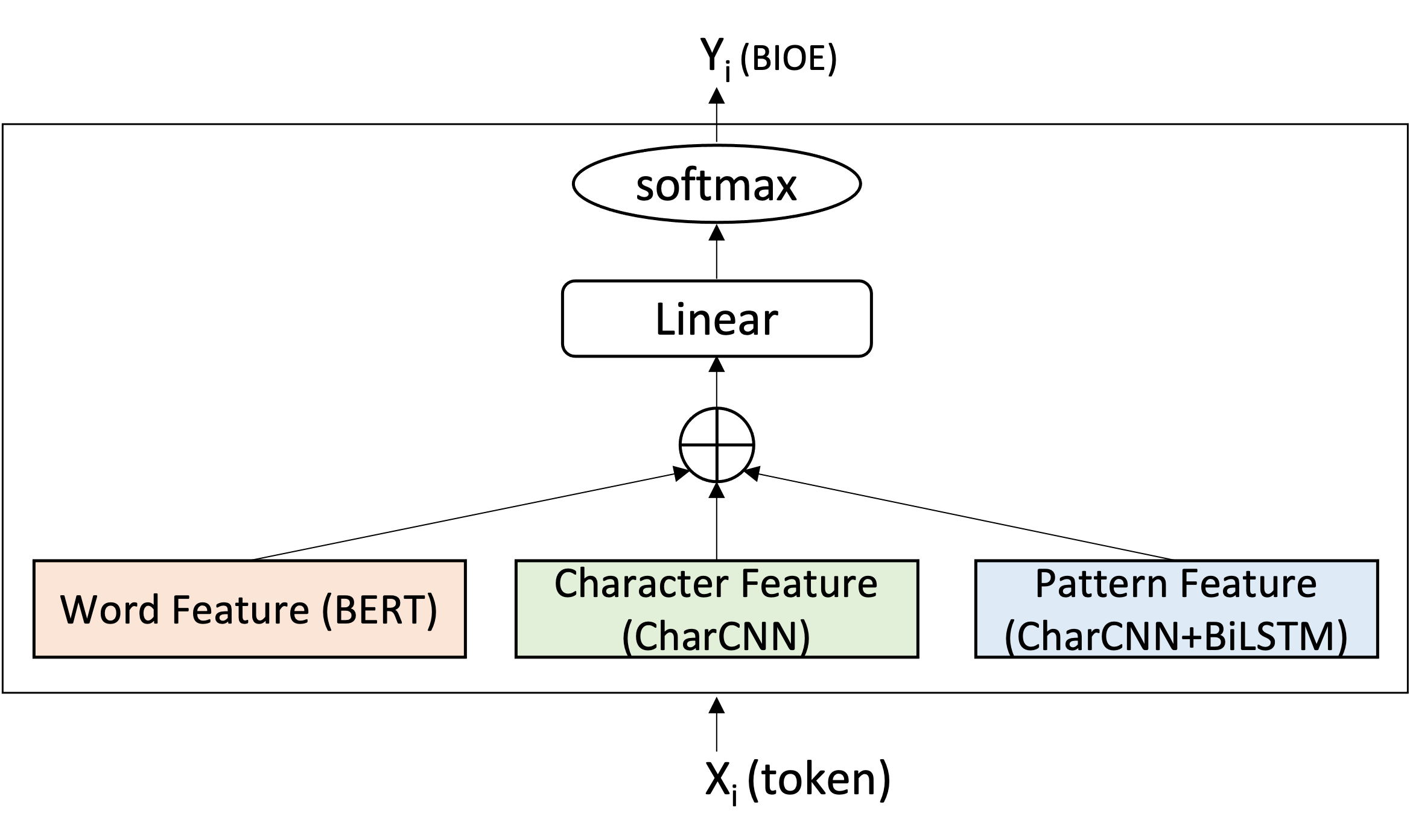}
   \caption{Token-level Schematic Model for \spandetect{}}
   \label{fig:token-span}
\end{figure}

A well-known problem in pipeline systems is error propagation. Inaccurate mention boundaries will lead to incorrect entity type classification. 
We observed that such boundary detection errors happen mostly for domain-specific terms which occur rarely and do not have a good semantic representation in the underlying BERT model. 
However, these domain specific terms often share patterns at character-level (e.g., chemical formulas). 
Thus we add character sequences and intrinsic orthographic patterns as additional features along with the BERT embeddings. 
The character and pattern features are shown to produce better word representations~\cite{carreras-etal-2002,limsopatham-collier-2016-bidirectional,boukkouri2020characterbert,LangeASK21}.

\paragraph{Character Sequence Feature} 
\label{sec:char-feature}
To learn character-level representation of each token, we use five one-dimensional CNNs with kernel sizes from $1$ to $5$, each having $16$ filters and $50$ input channels. 
Each token output from WordPiece Tokenizer is fed to the five CNN models simultaneously, which produce a $50$-dimensional embedding for each character.
These are max-pooled and the outputs from the CNNs are concatenated and passed through a linear layer with ReLU activation to get a $768$-dimensional character-level representation of the token. 
Figure~\ref{fig:char-feature} shows the process. 

\paragraph{Orthographic Pattern Feature} 
\label{sec:pattern-feature}
To capture the intrinsic orthographic patterns (or word shapes) of entity mentions at the sub-word level, 
we map all uppercase tokens to a single character, \texttt{U}, all lowercase tokens to \texttt{L}, all digit tokens to \texttt{D}.  
If a token contains a mix of uppercase, lowercase and digits, we map each lowercase character to \texttt{l}, uppercase to \texttt{u} and digit to \texttt{d}. Special characters are retained and BERT's special tokens, ``[CLS]'' and ``[SEP]'', are mapped to \texttt{C} and \texttt{S} respectively. 

We use 3 CNNs with the same setup as character sequence with kernel sizes of $1$ to $3$. 
Note that a contextual learning layer is needed to capture patterns in mentions spanning multiple tokens.
Thus, we pass the pattern-level embeddings for all tokens to a bidirectional LSTM with $256$ hidden dimensions as shown in Figure~\ref{fig:pattern-feature}.
%
Finally, the character and pattern features are concatenated with the BERT output for the token and fed to a final classifier layer as shown in  Figure~\ref{fig:token-span}.

\subsection{Span Classification}
\label{sec:class}
Given a sentence $\mathcal{S}=\langle w_1, w_2 \ldots w_n \rangle$ and a span $\langle s, e\rangle$,
this step determines the entity type 
for the span. 
Existing QA-based NER methods take the target entity type as the question (e.g.,  \textit{``Where is \texttt{person}?}) and return the corresponding mentions in the sentence.  
On the contrary, our model takes a mention as the question (e.g.,  \textit{``What is \texttt{Emily}?}) and outputs its entity type. 

During training, we create a training sample for each labeled entity mention in a sentence. 
During inference, the model gets the mention spans from \spandetect{} as its input. 
An input sample is created by appending the mention span text as \textit{``What is [mention]?"} to the input sentence (see top diagrams in Figure~\ref{fig:framework} for example).
This is fed to a BERT model and the pooled sequence embedding is fed to a fully connected layer and converted into a probability distribution over the entity types. 

\if
We note that many real world datasets have a class imbalance problem which may bias the model to choose more common entity types. 
A recent study \cite{li2019dice} has shown that using dice coefficient as the loss function is more beneficial than cross entropy loss in such situations. 
Inspired by this work, we trained the \spanclass{} using dice loss and show its effectiveness 
 through experiments and ablations later. 
\fi



\section{Experimental Results}
\label{sec:exp}
We demonstrate the effectiveness of our method in terms of performance and latency. 

\subsection{Datasets}
Table \ref{tab:datasets_summary} shows our datasets, including
three public benchmark datasets, \data{BioNLP13CG} \cite{pyysalo2015overview}, \data{OntoNotes5.0} \cite{weischedel2013ontonotes}, and \data{WNUT17} \cite{derczynski2017results},
 and a private dataset\footnote{The dataset curation procedure, entity types and their distribution is described in detail in Appendix \ref{sec:cti_dataset}.} (\securitydata{}) from the cybersecurity domain which contains news articles and technical reports related to malware and security threats.
%
These datasets cover not only the traditional whole-word entities like \type{Person} but also entity types with non-word mentions (e.g., chemical formulas) and very long mentions (e.g., URLs). 
\begin{table}[htb]
\setlength{\tabcolsep}{4pt}
\centering
\begin{small}
\begin{tabular}{ccccrrr}\toprule
\header{Dataset}    & \header{Type} &  \header{Density} & \header{Train} & \header{Dev} & \header{Test} \\  \midrule
\data{BioNLP13CG}   & $16$ & 3.59 &  $3,033$  & $1,003$ & $1,906$ \\
\securitydata & $8$  & 0.63 &  $38,721$ & $6,322$ & $9,837$\\
\data{OntoNotes5.0} & $18$ & 1.36 &  $59,924$ & $8,528$ & $8,262$\\  
\data{WNUT17}       & $6$  & 0.68 &  $3,394$  & $1,009$ & $1,287$\\
\bottomrule
\end{tabular}
\caption{Dataset Summary. \header{Type} and \header{Density} show number of entity types and average number of mentions per sentence.
\header{Train}, \header{Dev}, \header{Test} show number of sentences.
}
\label{tab:datasets_summary}
\end{small}
\end{table}

 \begin{table*}[h!]
\centering
\begin{small}
\begin{tabular}{ccccc}\toprule
Model & \data{BioNLP13CG} & \securitydata{} & \data{OntoNotes5.0} & \data{WNUT17} \\ \toprule
\our     & $86.75$ & $\mathbf{74.96}$ & $\mathbf{90.86}$ & $\textbf{57.25}$  \\
\nocharpattern{} & $86.70$ & $74.05$ & $90.58$ & $56.24$  \\  
\spanseq & $86.08$ & $73.84$ & $90.30$ & $56.10$ \\ \midrule
\bertqa & $86.68$ & $71.70$ & $89.02$  & $43.45$ \\
\bertseq & $\textbf{87.08}$ & $72.36$ & $88.64$ & $44.97$\\
\bottomrule
\end{tabular}
\caption{NER Performance Comparison (mention-level F1). \our{} is our proposed method. 
}
\label{tab:main}
\end{small}
\end{table*}

\begin{table*}[ht]
\centering
\begin{small}
\begin{tabular}{crrrrrrrr}\toprule
 Model & \multicolumn{2}{c}{\data{BioNLP13CG}} & \multicolumn{2}{c}{\securitydata{}} & \multicolumn{2}{c}{\data{OntoNotes5.0}} & \multicolumn{2}{c}{\data{WNUT17}} \\ \cmidrule{2-9}
    & Train & Inf.  & Train & Inf. & Train & Inf. & Train & Inf.  \\ \toprule
\our     & 241.2 & 57.7 & \textbf{1,455.7} & 120.0 & \textbf{3,007.8} & 183.0 & 122.9 & 26.0 \\
\bertqa & 1,372.8&323.3 & 8,771.0 &551.6 &  73,818.4 & 2,227.8  & 568.2 &91.2\\
\bertseq & \textbf{102.2} & \textbf{25.2} & 6,425.9 & \textbf{86.4} &  9,181.1 & \textbf{105.0}  & \textbf{101.3} & \textbf{18.6} \\
\bottomrule
\end{tabular}
\caption{Comparison of training and inference (Inf.) latency in seconds.}
\label{tab:train_time_ablation}
\end{small}
\end{table*}

\subsection{Experimental Setup}
We implement our baselines and our proposed system, \splitner{} in \deptool{pytorch} using \deptool{transformers}~\cite{transformers}. All models are trained on \textit{Nvidia Tesla V100} GPUs and use BERT-base architecture.
We use pretrained RoBERTa-\textit{base} \cite{liu2019roberta} backbone for all experiments with \data{OntoNotes5.0} corpus following \citet{ye2022packed, zhu2022boundary} and use SciBERT-\textit{scivocab-uncased} \cite{beltagy-etal-2019-scibert} for \data{BioNLP13CG} since this dataset has chemical formulas and scientific entities\footnote{We also experimented with BioBERT\cite{lee2020biobert} (\texttt{dmis-lab/biobert-base-cased-v1.1}) which gives similar trends. But SciBERT outperforms in our experiments.}. For \data{WNUT17}\footnote{BERTweet\cite{nguyen2020bertweet} model can also be used for \data{WNUT17}. 
We expect it to give same trends with even better performance figures.} and \securitydata{}, we use BERT-\textit{base-uncased} \cite{devlin2019bert}. Note that our model is a general two-step NER framework which has the performance benefits of QA-based and span-based approaches with efficiency. It can work with any BERT-based pretrained backbones.

The training data is randomly shuffled, and a batch size of $16$ is used with post-padding. The maximum sequence length is set to $512$ for \data{OntoNotes5.0}\footnote{Sentences in \data{OntoNotes5.0} are found to be longer and with maximum sequence length set to 256, lots of sentences get truncated. Hence we select a larger limit of 512.} and to $256$ for all other datasets.
For model optimization, we use cross entropy loss for span detection and dice loss\cite{li2019dice} for span classification.
All other training parameters are set to defaults in \deptool{transformers}. 
%


\subsection{Performance Evaluation}
We compare our method \our{} with the following baselines and variants. (1) \bertseq{}: The standard single-model sequence tagging NER setup which classifies each token using \texttt{BIOE} scheme. (2) \bertqa{}: The standard single-model QA-based setup which prefixes input sentences with a question describing the target entity type (e.g., \textit{Where is the \texttt{person} mentioned in the text?}); (3) \spanseq{}: A variant of our model which uses sequence tagging for span detection with our QA-based \spanclass{}; 
(4) \nocharpattern{}: This model is the same as our method but without the additional character and pattern features.
%
All other baselines use character and pattern features for fair comparison.
 We trained all models with 5 random seeds and report the mean mention-level Micro-F1 score in Table \ref{tab:main}. As can be seen, \our{} outperforms all baselines on three cross-domain datasets and gives comparable results on \data{BioNLP13CG}. We present further ablation studies on individual components of our system in Appendix \ref{sec:ablations} and a qualitative study in Appendix \ref{sec:quality}.

\subsection{Latency Evaluation}

We compare the latency of our method, \our{} and the two single-model NER methods.
Table \ref{tab:train_time_ablation} shows the training and inference times. 
Training time is measured for one epoch and averaged over 10 runs.
For a fair comparison, we report the training latency for our system as the sum of span detection and classification even though they can be trained in parallel. 

The results show that, compared to \bertqa{}, our method is 5 to 25 times faster for training and about 5 times faster for inference, and it is especially beneficial for large datasets with many entity types.
Compared to \bertseq{}, our method is slightly slower but achieves much better F1 scores (Table~\ref{tab:main}). 
These results validate \our{}'s effectiveness in achieving the balance between performance and time efficiency.

\section{Related Work}
\label{sec:related}
In recent years, deep learning has been increasingly applied for NER \cite{torfi2020natural, li2020survey}, a popular architecture being CNN-LSTM-CRF \cite{ma2016end,xu2021better} and BERT \cite{devlin2019bert}. \citet{li2020MRC,li2019dice} propose a QA-based setup for NER using one model for both span detection and classification. \citet{li2020MRC,Jiang20,Ouchi20,fu2021spanner,zhu2022boundary} perform NER as a span prediction task. However, they enumerate all possible spans in a sentence leading to quadratic complexity w.r.t. sentence length. Our model does a token-level classification and hence is linear.

\citet{xu2021better} propose a Syn-LSTM setup leveraging dependency tree structure with pretrained BERT embeddings for NER. \citet{yan2021unified} propose a generative framework leveraging BART \cite{lewis2019bart} for NER. \citet{yu2020named} propose a biaffine model utilizing pretrained BERT and FastText \cite{bojanowski2017enriching} embeddings along with character-level CNN setup over a Bi-LSTM architecture. All of these models report good performance on \data{OntoNotes5.0}, however, using \bertlarge{} architecture. \citet{nguyen2020bertweet} propose the BERTweet model by training BERT on a corpus of English tweets and report good performance on \data{WNUT17}. \citet{wang2021improving} leverage external knowledge and a cooperative learning setup.
On \data{BioNLP13CG}, \citet{crichton2017neural} report $78.90$ F1 in a multi-task learning setup and \citet{neumann2019scispacy} report $77.60$ using the SciSpacy system. \our{} outperforms both of these by a large margin.

\section{Conclusion}
Using the QA-framework for both span detection and span classification, we show that this division of labor is not only effective but also significantly efficient through experiments on multiple cross-domain datasets. 
Through this work, we open up the possibility of breaking down other complex NLP tasks into smaller sub-tasks and fine-tuning large pretrained language models for each task. 

\if
Nevertheless, our approach has a pipeline-based structure and hence errors made during the span detection step propagate to the span classification stage, unlike other single model-based NER systems.
However, our breakdown allows more flexibility to optimize the models for the individual sub-tasks more effectively.
We show that explicitly modeling character and pattern features helps to detect span boundaries better,
 and dice loss helps to handle class imbalance issues.
Through these customized optimizations, we can achieve better model efficiency and effectiveness.

Interestingly, our span classification setup is the reverse of a standard QA-setup for NER. 
For a sentence, \textit{Emily lives in United States}, rather than asking a question like, \textit{``What is the \texttt{person} in the text?"} as done in other QA-based NER systems, we ask \textit{``What is \texttt{Emily}?"}. 
This opens up prospects for more intuitive and creative approaches for NER and shows the power of BERT model yet again in capturing the semantics and query intent well.

During our qualitative study, we identify that span detection serves as a primary bottleneck for  the overall NER performance 
 and encourage the research community to design architectures or new training objectives to detect mention boundaries more effectively.    
Currently, in our \textit{Span Detection Module}, all entity mentions are grouped into a single class. As a potential future work, we expect to get even better performance by a hierarchical extension of our setup. At the top level, we detect mentions belonging to some crude categories and gradually break them down into more fine-grained categories. 
\fi

\section*{Limitations}
Our proposed approach requires to train two independent classification models. 
While the models can be trained in parallel, 
this requires larger GPU memory. 
For the experiments, we trained two BERT-base models, which have around 220M trainable parameters when trained in parallel. This requires almost twice the GPU memory compared to a single BERT-base NER model, having around 110M trainable parameters.

Owing to a pipeline-based structure, the overall performance of our system is upper bounded by the performance of \spandetect{} which has lots of potential for improvement. On dev set, we find that around 30\% of errors for \data{OntoNotes5.0} and \data{BioNLP13CG}, and around 22\% errors on \data{WNUT17} are just due to minor boundary detection issues. Their entity types are being detected correctly. We henceforth encourage the research community to design architectures or new training objectives to detect mention boundaries more effectively. Currently, in our \spandetect{}, all entity mentions are grouped into a single class. As a potential future work, we expect to get even better performance by a hierarchical extension of our setup. At the top level, we can detect mentions belonging to some crude categories and gradually break them down into more fine-grained categories. 

\section*{Acknowledgements}
This work was conducted under the IBM-Illinois Center for Cognitive Computing Systems Research (C3SR\footnote{https://www.c3sr.com}), while the first author was a graduate student at University of Illinois Urbana-Champaign and an intern at IBM. We are also very grateful to Prof. Jiawei Han, University of Illinois Urbana-Champaign, for his continued guidance, support and valuable feedback throughout this work.

\bibliography{main}
\bibliographystyle{acl_natbib}

\appendix

\section{Performance Ablations}
\label{sec:ablations}

\begin{table*}[t!]
\centering
\begin{small}
\begin{tabular}{ccccccccccccc}\toprule
\multirow{2}{*}{\shortstack{Span Detection\\Features}} & \multicolumn{3}{c}{\data{BioNLP13CG}} & \multicolumn{3}{c}{\securitydata{}} & \multicolumn{3}{c}{\data{OntoNotes5.0}} & \multicolumn{3}{c}{\data{WNUT17}} \\ \cmidrule{2-13}
 & P & R & F1 & P & R & F1 & P & R & F1 & P & R & F1 \\ \midrule
\method{+CharPattern} & \textbf{91.43} & 90.70 & \textbf{91.06} & \textbf{80.59} & 77.21 & \textbf{78.86} & \textbf{92.17} & \textbf{92.83} & \textbf{92.50} & \textbf{73.38} & \textbf{44.25} & \textbf{55.21}  \\
\method{-CharPattern} & 90.31 & \textbf{91.03} & 90.67 &79.65 & \textbf{77.77} & 78.70 & 91.96 & 92.79 & 92.37 & 72.63 & 44.06 &54.85  \\
\bottomrule
\end{tabular}
\caption{\spandetect{} performance with and without character and pattern features.
}
\label{tab:det_ablation}
\end{small}
\end{table*}

Here, we study the individual components of our system, \our{} in detail.
First, we investigate the effectiveness of the additional character and pattern features for span detection. 
As we can see from Table~\ref{tab:det_ablation}, the character and pattern features improve the NER performance for all datasets.

We also study the effect of the character and pattern features separately. 
Table~\ref{tab:feature_ablation} shows this ablation study on the \data{BioNLP13CG} dataset. 
As we can see, adding the character feature or the pattern feature alone makes a small change in the performance.
Interestingly, the character feature helps with recall, while the pattern features improves precision, and, thus, 
adding them together improves both precision and recall.
However, adding part-of-speech (POS) in addition to the character and pattern features shows little impact on the performance.

\begin{table}[h!]
\centering
\begin{small}
\begin{tabular}{lccc}\toprule
Features & P & R & F1 \\ \midrule
\texttt{Base Model} & 90.31 &   91.03 &  90.67 \\
\texttt{+Char} & 89.85 &   91.45 &  90.64 \\
\texttt{+Pattern} & 91.29 &   90.22 & 90.75 \\
\texttt{+Char+Pattern} & \textbf{91.43} & \textbf{90.70} & \textbf{91.06} \\
\texttt{+Char+Pattern+POS} & 91.14 & 90.64 & 90.89 \\
\bottomrule
\end{tabular}
\caption{\spandetect{} performance for  \data{BioNLP13CG} with different feature sets. 
\texttt{Base Model} does not use character and pattern features. 
}
\label{tab:feature_ablation}
\end{small}
\end{table}

%
Next, we compare dice loss and cross-entropy loss for their effectiveness in handling the class imbalance issue in span classification. 
As shown in Table~\ref{tab:class_ablation}, dice loss works better for imbalanced data confirming the results found in \citet{li2019dice}.

\begin{table}[h!]
\centering
\begin{small}
\begin{tabular}{ccc}\toprule
Dataset & Dice Loss &   Cross Entropy Loss \\ \toprule 
\data{BioNLP13CG} & \textbf{94.27}  & 94.04\\
\data{CyberThreats} & \textbf{87.84} & 87.58\\
\data{OntoNotes5.0} &  \textbf{96.74}  & 96.50 \\
\data{WNUT17} & \textbf{73.40}    & 73.31  \\
\bottomrule
\end{tabular}
\caption{Span classification performance comparison}
\label{tab:class_ablation}
\end{small}
\end{table}
Finally, we experimented with different question sentences in \spandetect{} to check if BERT is giving any importance to the query part.
As shown in Table~\ref{tab:question_ablation}, different queries do have a minor impact but as expected, the model mostly learns not to focus on the query part as can be seen by the comparable results with \textit{<empty>} query. 
\begin{table}[h!]
\centering
\begin{small}
\begin{tabular}{p{2.45in}c}\toprule
Question Type &  F1 \\ \midrule
\scriptsize{\textit{\textbf{Extract important entity spans from the following text.}}} &  \textbf{90.67} \\
\textit{Where is the entity mentioned in the text?} &   90.38 \\
\textit{Find named entities in the following text.} &  90.32 \\
\textit{<empty>} &   90.48 \\
\bottomrule
\end{tabular}
\caption{\spandetect{} performance for  \data{BioNLP13CG} using different questions. <empty> denotes an empty question sentence.
All experiments were done using \texttt{Base Model} in Table~\ref{tab:feature_ablation}.}
\label{tab:question_ablation}
\end{small}
\end{table}

\subsection{Discussions}
\label{subsec:discussion}
From the results of the experiments described in Section~\ref{sec:exp} together with the ablation studies, we make the following observations:

\begin{itemize}[topsep=2pt,itemsep=1pt,parsep=1pt]
    \item As shown in Table \ref{tab:main}, \our{} outperforms both the sequence tagging and QA-based baselines on three cross-domain datasets and performs on-par on \data{BioNLP13CG}.
    
     \item The division of labor allows each model to be optimized for its own sub-task.       Adding character and pattern features improves the accuracy of \spandetect{} (Table \ref{tab:det_ablation}). 
     However, adding these same features in \spanclass{} was found to deteriorate the performance.
    Similarly,  dice loss improves the performance for \spanclass{} (Table \ref{tab:class_ablation}), 
    but no such impact was observed for \spandetect{}.
    
    \item Span detection using the QA setting  is slightly more effective than  the sequence tagging setup as done in \spanseq{} (Table \ref{tab:main}).

    \item Our model has more representative power than the baseline approaches, because it leverages two BERT models, each working on their own sub-tasks. 
    
    \item  It also leverages the QA framework much more efficiently than the standard single-model QA system (Table \ref{tab:train_time_ablation}). The margin of improvement is more pronounced when the data size and number of entity types increase.
    
    \item The training time for our model in Table \ref{tab:train_time_ablation} considers \spandetect{} and \spanclass{}  being trained sequentially. However, the two components can be trained in parallel, reducing the overall train time significantly. The sequential execution is necessary only at inference time.
    
     \item \data{WNUT17} has a diverse range of rare and emerging entities crudely categorized into $6$ entity types. A single-model NER system may get confused and try to learn sub-optimal entity-specific extraction rules. Our task segregation allows \spandetect{} to form generalized extraction rules which is found to be more effective as shown in Table \ref{tab:main}.
    
     \item As a sidenote, all the models built in this work outperform the previously published approaches on \data{BioNLP13CG} (Table \ref{tab:main}), thus setting new state-of-the-art results. The credit goes to the SciBERT model
    and the additional character and pattern features.
    
    
    
\end{itemize}

\section{Qualitative Analysis}
\label{sec:quality}
Table \ref{tab:quality} shows some sample predictions by our method, \our{} and compares them with our single-model NER baseline, \bertqa{}. 
From the results, we observe that:
\begin{itemize}[topsep=0pt,itemsep=1pt,parsep=1pt]
    \item \our{} is better in detecting emerging entities and out-of-vocabulary (OOV) terms (e.g., movie titles and softwares). 
    This can be attributed to \spandetect{} being stronger in generalizing and sharing entity extraction rules across multiple entity types.
    
    \item \bertqa{} gets confused when entities have special symbols within them (e.g., hyphens and commas). Our character and orthographic pattern features  help handle such cases well.
    
    \item \bertqa{} model develops a bias towards more common entity types (e.g., \type{PERSON}) and misclassifies rare entity mentions 
    when they occur in a similar context. \our{} handles such cases well thanks to the dedicated \spanclass{} using dice loss.
\end{itemize}

\begin{table}[ht]
\setlength{\tabcolsep}{2pt}
\centering
\begin{small}
\begin{tabular}{ll}\toprule
Category &  {~~~~~~}Example Sentence \\\toprule
\multirow{2}{*}{\shortstack{General \\ Detection}}     
     & CVS selling their own version of ... \\
     & \textit{\textcolor{red}{CVS}} selling their own version of ... \\ \midrule
\multirow{2}{*}{\shortstack{Emerging \\ Entities}}  
    & {\footnotesize Rogue One create a plot hole in Return of the Jedi}  \\
     & {\footnotesize \textit{\textcolor{blue}{Rogue One}} create a plot hole in \textit{\textcolor{blue}{Return of the Jedi}}}  \\ \midrule
\multirow{2}{*}{\shortstack{Scientific \\ Terms}} 
     & Treating EU - 6 with anti-survivin antisense ...  \\
     &\ Treating \textit{\textcolor{brown}{EU - 6}} with anti-survivin antisense ... \\ \midrule
\multirow{2}{*}{Boundary}  
   & Hotel Housekeepers Needed in Spring , \textit{\textcolor{green}{TX}} ...  \\
    & Hotel Housekeepers Needed in \textit{\textcolor{green}{Spring , TX}} ...  \\ \midrule
\multirow{2}{*}{\shortstack{OOV \\ Terms}}  
   & Store SQL database credentials in a webserver \\
    & Store \textit{\textcolor{magenta}{SQL}} database credentials in a webserver \\ \midrule
\multirow{2}{*}{\shortstack{Entity \\ Type}}  
  &  Why do so many kids in \textit{\textcolor{green}{Digimon}} wear gloves? \\
    &  Why do so many kids in \textit{\textcolor{magenta}{Digimon}} wear gloves? \\ \bottomrule
\end{tabular}
\if false
\begin{tabular}{ll}\toprule
\hspace{-3mm} Category &  Example Sentence \\\toprule
\hspace{-3mm}\multirow{2}{*}{\shortstack{General \\ Detection}}     
     & \hspace{-2mm}CVS selling their own version of ... \\
     & \hspace{-2mm}\textit{\textcolor{red}{CVS}} selling their own version of ... \\ \midrule
\hspace{-3mm}\multirow{2}{*}{\shortstack{Emerging \\ Entities}}  
    & \hspace{-2mm}{\footnotesize Rogue One create a plot hole in Return of the Jedi}  \\
     & \hspace{-2mm}{\footnotesize \textit{\textcolor{blue}{Rogue One}} create a plot hole in \textit{\textcolor{blue}{Return of the Jedi}}}  \\ \midrule
\hspace{-3mm}\multirow{2}{*}{\shortstack{Scientific \\ Terms}} 
     & \hspace{-2mm}Treating EU - 6 with anti-survivin antisense ...  \\
     &\hspace{-2mm} Treating \textit{\textcolor{brown}{EU - 6}} with anti-survivin antisense ... \\ \midrule
\hspace{-3mm}\multirow{2}{*}{Boundary}  
   & \hspace{-2mm}Hotel Housekeepers Needed in Spring , \textit{\textcolor{green}{TX}} ...  \\
    & \hspace{-2mm}Hotel Housekeepers Needed in \textit{\textcolor{green}{Spring , TX}} ...  \\ \midrule
\hspace{-3mm}\multirow{2}{*}{\shortstack{OOV \\ Terms}}  
   & \hspace{-2mm}Store SQL database credentials in a webserver \\
    & \hspace{-2mm}Store \textit{\textcolor{magenta}{SQL}} database credentials in a webserver \\ \midrule
 \hspace{-3mm}\multirow{2}{*}{\shortstack{Entity \\ Type}}  
  & \hspace{-2mm} Why do so many kids in \textit{\textcolor{green}{Digimon}} wear gloves? \\
    & \hspace{-2mm} Why do so many kids in \textit{\textcolor{magenta}{Digimon}} wear gloves? \\ \bottomrule
\end{tabular}
\fi
\caption{Qualitative comparison of \our{} and \bertqa{} systems. For each category, the first line shows the result of \bertqa{},
and the second line shows the result of \our{}. 
The words in italics are the entity mentions extracted by the systems color-coded as \textcolor{red}{\type{Org}}, \textcolor{blue}{\type{Creative Work}}, \textcolor{brown}{\type{Gene}}, \textcolor{green}{\type{Location}} and \textcolor{magenta}{\type{Product}}.
}
    \label{tab:quality}
\end{small}
\end{table}

\section{\securitydata{} Dataset}
\label{sec:cti_dataset}
  
  The \securitydata{} dataset is curated from a collection of 967 documents which include cybersecurity news articles and white papers published online by reputable companies and domain knowledge experts.
 These documents usually provide deep analysis on a certain malware, a hacking group or a newly discovered vulnerability (like a bug in software that can be exploited).
   The documents were published between 2016 and 2018. 
  We split the dataset into the train, development, and test sets as shown in Table~\ref{tab:data}. 

A team of cybersecurity domain experts labeled the dataset for the following 8 entity types.   These types were selected based on the STIX (Structured Threat Information Expression) schema which is used to exchange cyber threat intelligence. For more detailed information about the 8 types, please refer the STIX documentation\footnote{\url{https://stixproject.github.io/releases/1.2}}.
 \begin{itemize} [topsep=2pt,itemsep=1pt,parsep=1pt]
     \item CAMPAIGN: Names of cyber campaigns that  describe a set of malicious activities or attacks  over a period of time.
     \item COURSE OF ACTION: Tools or actions to take in response to cyber attacks. 
     \item EXPLOIT TARGET: Vulnerabilities that are targeted for exploitation.
     \item IDENTITY: Individuals, groups or organizations.
     \item INDICATOR: Objects that are used to detect suspicious or malicious cyber activity such as domain name, IP address and file names.
     \item MALWARE: Names of malicious codes used in cyber crimes.
     \item RESOURCE: Tools that are used in cyber attacks.
     \item THREAT ACTOR: Individuals or groups that commit cyber crimes.
 \end{itemize}

\begin{table}
\centering
\begin{small}
\begin{tabular}{lcccc} \toprule
                  & Train & Test  &  Dev & Total\\ \midrule
  \# documents & 667  & 133 & 167 & 967 \\
  \# sentences &  38,721  & 9,837 & 6,322 & 54,880 \\
  \# tokens    &  465,826  & 119,613 & 92,788 & 678,227 \\
  \bottomrule
\end{tabular}
\end{small}
\caption{Summary of the \securitydata{} corpus showing the number of documents, sentences and tokens in each dataset.}
\label{tab:data}
\end{table}

 Table~\ref{tab:entity_stat} and Table~\ref{tab:entity} show
  the statistics of the entity types in the corpus  and some sample mentions of these types respectively.

\makeatletter
\setlength{\@fptop}{20pt}
\makeatother

 \begin{table}[ht!]
\centering
\begin{small}
\begin{tabular}{crrr} \toprule
Entity Type & Train & Dev & Test \\ \midrule
CAMPAIGN & 247 & 27 & 85\\
COURSE OF ACTION & 1,938 & 779 & 329\\
EXPLOIT TARGET & 5,839 & 1,412 & 1,282\\
IDENTITY & 6,175 & 1,262 & 1,692\\
INDICATOR & 3,718 & 1,071 & 886 \\
MALWARE & 4,252 & 776 & 1,027 \\
RESOURCE & 438 & 91 & 114 \\
THREAT ACTOR & 755 & 91 & 144 \\
\bottomrule
\end{tabular}
\end{small}
\caption{The number of mentions for each entity type in the train, development and test sets}
\label{tab:entity_stat}
\end{table}

 \begin{table}[ht!]
\centering
\begin{footnotesize}
\begin{tabular}{p{0.6in}p{2.2in}}
\toprule
Entity Type &  {~~~~~~~~~~~~~~~~~~~~~}Examples \\ \toprule
\multirow{3}{*}{CAMPAIGN}  &  ``Operation Pawn Storm'', ``The Mask''\\
    & ``MiniDuke'', ``Woolen-Goldfish''\\ & ``Ke3chang'' \\ \midrule
\multirow{6}{*}{\shortstack{COURSE \\ OF \\ ACTION}}  &  ``Trojan.Poweliks Removal Tool'' \\  & ``HPSBHF03535'',  ``TDSSKiller''\\
  & ``cd\  chktrust -i FixTool.exe'' \\
  & ``http://www.ubuntu.com/usn/usn-2428-1'' \\
  &``Initial Rapid Release version June 15,\\
  &2015 revision 02'' \\
    \midrule
\multirow{3}{*}{\shortstack{EXPLOIT \\ TARGET}}  & ``CVE-2015-8431'', ``Adobe Flash Player''\\
 & ''Ubuntu'', ``Windows'', ``CGI.pm'' \\
 & ``version 20.0.0.306 and earlier'' \\ \midrule
\multirow{2}{*}{IDENTITY}  & ``Symantec'', ``Jon DiMaggio'', ``Belgium'' \\
  & ``Kaspersky Lab'', ``RSA''\\ \midrule
\multirow{3}{*}{INDICATOR} & ``C:\textbackslash{}WINDOWS\textbackslash{}assembly\textbackslash{}GAC\_MSIL'' \\   & ``hxxp://deschatz-army.net'', ``67.23.112.226''\\ & ``b4b483eb0d25fa3a9ec589eb11467ab8'' \\ \midrule
\multirow{4}{*}{MALWARE} & ``ChewBacca'', ``SONAR.AM.E.J!g13'' \\
 & ``Trojan.Poweliks'', ``BlackHole'' \\ & ``TDL3'', ``LockyZeus'' \\
 & ``JS/TrojanDownloader.Nemucod'' \\
 \midrule
\multirow{3}{*}{RESOURCE} & ``IRC'', ``Tor'', ``DroidPlugin'', ``Onion'' \\  
  & ``PowerShell'', ``Google Play'' \\
 &``Free Pascal 2.7.1.'', ``Teamviewer'' \\ \midrule
\multirow{3}{*}{\shortstack{THREAT \\ ACTOR}}  & ``ProjectSauron'', ``Strider'', ``Ogundokun'' \\
  & ``APT28'', ``APT 28'', ``Fancy Bear'' \\
  & ``Pro\_Mast3r'', ``Equation Group'' \\ 
\bottomrule
\end{tabular}
\end{footnotesize}
\caption{Sample entity mentions for each type in the \securitydata{} corpus}
\label{tab:entity}
\end{table}

\end{document}